\documentclass[11pt]{article}
\usepackage[margin=1in]{geometry}
\usepackage{amsmath,amssymb,amsfonts,bbm, mathrsfs}
\usepackage{graphicx}
\usepackage{natbib}
\usepackage{setspace}
\usepackage{algorithmic}
\usepackage{algorithm}
\usepackage{subcaption}
\newcommand{\Bta}{\mathrm{Beta}}

\newcommand{\zap}[1]{}

\usepackage{natbib}
 \bibpunct[, ]{(}{)}{,}{a}{}{,}%

\begin{document}
\title{A Markov Decision Process Analysis of the Cold Start Problem in Bayesian Information Filtering}
\author{
Xiaoting Zhao, Peter I. Frazier \\
School of Operations Research and Information Engineering\\
Cornell University,
Ithaca, NY 14850 \\
\texttt{xzhao@cornell.edu, pf98@cornell.edu}
}
\date{\today}
\maketitle
\begin{abstract}
We consider the information filtering problem, in which we face a stream of items, and must decide which ones to forward to a user to maximize the number of relevant items shown, minus a penalty for each irrelevant item shown.  Forwarding decisions are made separately in a personalized way for each user. We focus on the cold-start setting for this problem, in which we have limited historical data on the user's preferences, and must rely on feedback from forwarded articles to learn which the fraction of items relevant to the user in each of several item categories.  Performing well in this setting requires trading exploration vs. exploitation, forwarding items that are likely to be irrelevant, to allow learning that will improve later performance. In a Bayesian setting, and using Markov decision processes, we show how the Bayes-optimal forwarding algorithm can be computed efficiently when the user will examine each forwarded article, and how an upper bound on the Bayes-optimal procedure and a heuristic index policy can be obtained for the setting when the user will examine only a limited number of forwarded items. We present results from simulation experiments using parameters estimated using historical data from arXiv.org.
\end{abstract}

%% \subsection{Keywords for paper submission}
%% Your NIPS paper can be submitted with any of the following keywords (more than one keyword is possible for each paper):

%% \begin{verbatim}
%% Bioinformatics
%% Biological Vision
%% Brain Imaging and Brain Computer Interfacing
%% Clustering
%% Cognitive Science
%% Control and Reinforcement Learning
%% Dimensionality Reduction and Manifolds
%% Feature Selection
%% Gaussian Processes
%% Graphical Models
%% Hardware Technologies
%% Kernels
%% Learning Theory
%% Machine Vision
%% Margins and Boosting
%% Neural Networks
%% Neuroscience
%% Other Algorithms and Architectures
%% Other Applications
%% Semi-supervised Learning
%% Speech and Signal Processing
%% Text and Language Applications

%% \end{verbatim}

\section{Introduction}
We study the information filtering problem, in which a user faces a stream of time-sensitive items (emails, blog posts, scientific articles), some of which are interesting to the user, but many of which are uninteresting.  We wish to design an automatic system that automatically filters this stream, showing as many relevant items to the user as possible, while showing few irrelevant items.

When historical relevance data from the user is abundant, we can train a statistical classifier, and forward only those items predicted to be relevant. However, when historical data from the user is limited, e.g., because the user is new, or because we are dealing with new kinds of items, we face the {\it cold start problem}, in which we do not have enough training data to build a reliable classifier.

In this setting, it may be advantageous to {\it explore}, i.e., forward some items predicted to be irrelevant, just to learn their true relevance and improve future predictions. Too much exploration, however, will lead to forwarding many irrelevant items. Thus, an information filtering system should also put some weight on {\it exploitation}, i.e., forwarding only those items predicted to be relevant. 

We study this tradeoff between exploration vs. exploitation in a Bayesian setting, using a Markov decision process (MDP).  We show how the MDP defining the Bayes-optimal algorithm for making forwarding decisions may be solved efficiently using a decomposition, when irrelevant items are penalized by a user-specified cost per item shown.  We then show how to use this solution to provide a ranking over items when a cost-per-item is unknown not given and users are only willing to examine a limited number of items, but users examine each forwarded item. 

Exploration vs. exploitation has been studied extensively in the context of the multi-armed bandit problem in both Bayesian treatments \citep{GiJo74,Wh80,GittinsGlazebrookWeber}, and non-Bayesian treatments \citep{AuCeFrSc95,AuCeFi02}. This tradeoff between exploration and exploitation, which appears in other problem domains including reinforcement learning \citep{KaLiCa98,SuttonBartoRL98,AuerNearOptimalRegret2010}, approximate dynamic programming \citep{Powell04learningalgorithms,PowellADP2007}, revenue management \citep{ArCa10,BeZe09,den2013simultaneously}, optimization algorithms \citep{XieFrazier2013a,FrazierPowellDayanik2008}, and inventory control \citep{LaPo99,DiPuBi02}. In information retrieval problems, exploration vs. exploitation has also been studied in \cite{ZhXuCa03,AgarwalChenElango2009,YuBrKlJo09,Hofmann2013}.

We are motivated by a personalized information system we are building for the electronic repository of scientific articles, arXiv.org.  In popular categories like astro-ph (Astrophysics) and hep-th (high-energy physics), roughly 80 new articles are submitted each week \citep{arxiv}, which creates a challenge for scientists who wish to remain abreast of new arXiv articles directly relevant to their research.  Our experimental results use parameter settings estimated from historical data from arXiv.org.

This paper builds on the previous work \cite{ZhaoFrazier2014}, which considers an information filtering problem in a Bayesian setting, and uses dynamic programming to find the Bayes-optimal strategy for trading exploration and exploitation.  There are two main differences between the model considered in that paper, and the one considered in the current paper.
First, in \cite{ZhaoFrazier2014}, users provide {\it immediate} feedback on forwarded items, while in the current paper, we allow items to queue in the system until the next user visit.  It is only upon visiting the system that the user provides feedback.  This ``periodic review'' assumption is more realistic in many information filtering systems.Second, \cite{ZhaoFrazier2014} assumes that users provide a unit cost for forwarding to the information filtering system, while in the current paper we provide a method for ranking results that allows this cost to be unknown.

Our focus on a Bayesian setting and Bayes-optimal procedures (rather than procedures that are just optimal up to a constant), is in contrast with the portion of the literature on multi-armed bandits that examines regret in a worst-case setting, and provides algorithms that have optimal dependence on time or other problem parameters, but ignore constants.  This focus allows us to apply our method profitably in small-sample regimes, where the best worst-case guarantee would be much worse than the best average-case guarantee, and where constants are just as important as the dependence on time. A downside of our focus on the Bayesian setting is that it requires us to choose a prior distribution to use when measuring performance. However, in many applied settings, including arXiv.org, we argue that a reasonable prior distribution can be estimated from historical data.

In Section \ref{sec:filtering_model}, we formulate the information filtering problem with periodic reviews and a fixed unit cost for forwarding an item. In Section \ref{sec:LagrangeRelaxation}, we consider the case where the unit cost is unknown and there is a budget constraint on the total number of items that a user can view. We then show how to derive a ranking from this budgeted problem. Lastly, we show experimental results in Section \ref{sec:experiments}.

\section{Mathematical Model with a Unit Cost for Forwarding}
\label{sec:filtering_model}
We assume that items arrive to the system according to a Poisson process with rate $\lambda>0$. Each item is categorized into (exactly) one of $k$ categories, $\{1, ..., k\}$, and the category is observed as it enters the system. For systems without explicit categorization, the categories could be obtained by running a clustering algorithm on previous collected items in a pre-processing step.
\begin{figure}[h]
\begin{center}
\includegraphics[scale=0.28]{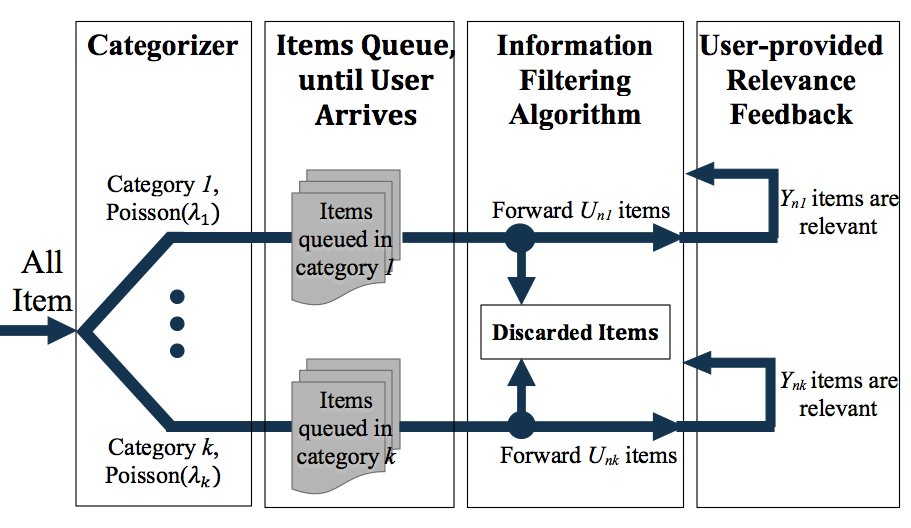}
\end{center}
\caption{Schematic of the information filtering problem with periodic reviews. }
\end{figure}
Let $X_i$ denote the category of the $i^{th}$ arriving item. We assume that the $X_i$ are independent and identically distributed, and we let $p_x=P(X_i=x)>0$.  Thus, items in each category $x \in \{1, ..., k\}$ arrive according to a Poisson process with rate $\lambda_x=p_x \lambda$. %In addition, let $(t_i: i=1,2,...)$ denote the arrival time of each item. 

Each category $x$ has some latent unobserved value $\theta_x \in [0, 1]$ measuring the probability that an item from category $x$ is relevant to the user. Let $\theta = [\theta_1, ..., \theta_k]$. We place a Bayesian prior distribution on this $\theta$, given by $\theta_x \sim \Bta(\alpha_{0x}, \beta_{0x})$, with independence across $x$, for some parameters $\alpha_{0x}$ and $\beta_{0x}$, typically estimated using historical data from (other) long-time users on older items. In our model, $\theta$ is assumed to stay static over the user's lifetime. 

The optimal tradeoff of exploration vs. exploitation will depend on how long the user interacts with our stream of items. Let $T$ be the length of time that the user uses our information filtering system.  $T$ is unknown a priori, and we model it as an exponential random variable with parameter $r$.

The previous model in \cite{ZhaoFrazier2014} provides a Bayes-optimal algorithm that analyzes the situation in which the user is always available to provide immediate feedback on each arriving item. However, in many real systems, users do not behave like this. Instead, they arrive periodically to review items that have queued in the system since their last visit.

Our model assumes that the user visits the system at time points separated by exponentially distributed inter-arrival times, which are independent and have parameter $s$. Let $N$ be the number of user visits before $T$. Here, we assume that at each visit, he or she examines all items forwarded from the stream since the last visit. Below, we study a problem variant in which the number of items the user is willing to examine on each visit is constrained.

At the $n^{th}$ user visit, the posterior on $\theta_x$ is $\Bta(\alpha_{nx}, \beta_{nx})$, for some $\alpha_{nx}$, $\beta_{nx}$. Based on $(\alpha_{nx}, \beta_{nx})$, we choose $U_{nx}$, denoting the maximum number of items to forward to the user from category $x$. For simplicity, we choose $U_{nx}$ before observing $L_{nx}$, which is the number of items queued in category $x$, then we show $Z_{nx}=\min(L_{nx}, U_{nx})$ items from category $x$ to the user in his or her $n^{th}$ visit. For computationally convenience, we require $U_{nx} \le M$, where $M < \infty$. 

With the decision $U_{nx}$ and forwarded $Z_{nx}$ items, the user provides explicit feedback, denoted by $Y_{nx}$, reflecting the actual relevance of the shown items to the user. Conditioning on $U_{nx}$, $\theta_x$ and $L_{nx}$, $Y_{xn}$ is binomial with a probability of $\theta_{x}$ being relevant to the user, that is,
$$Y_{nx}~|~ \theta_x, U_{nx}, L_{nx} \sim \text{Binomial}\left(Z_{nx}, \theta_x \right).$$
We also assume there is a unit cost, $c$, for forwarding each item to the user. Thus, we collect reward $Y_{nx}-cZ_{nx}$ in each step. We define a policy $\pi$ as a sequence of functions, $(\pi_1,  \pi_2, ..., \pi_N)$, where each $\pi_n: (\{0, 1..., M\}^{k} \times \mathbb{N}^k \times \{0, 1..., M\}^{k})^{n-1} \mapsto \{0, 1, ..., M\}^k$ maps history, $\{U_{\ell x}, L_{\ell x}, Y_{\ell x}: \ell \le n-1, x \in \{1,..., k\}\}$ into actions. Let $\Pi$ be the set of all such policies. Our objective becomes to find an optimal policy $\pi \in \Pi$ that maximizes total expected reward:
\begin{eqnarray}
\begin{split}
\label{eq:originalEq}
&\sup_{\pi \in \Pi} \mathbb{E}^\pi \left [ \sum^N_{n=1}\sum^k_{x=1}(Y_{nx} - c Z_{nx})\right].
\end{split}
\end{eqnarray}

\subsection{Solution and Computation Method}
Due to the independence assumption across $\theta_{x}$, we can decompose the original problem with $k$-categories into a sum of $k$ independent sub-problems, each of which can be solved via stochastic dynamic programming. Equation \eqref{eq:originalEq} is rewritten as,
$$\sup_{\pi \in \Pi} E^{\pi} \left[ \sum^k_{x=1} \sum^N_{n=1} (Y_{nx}-cZ_{nx}) \right] = \sum^k_{x=1} \sup_{\pi(x) \in \Pi(x)} E^{\pi(x)} \left[\sum^N_{n=1}(Y_{nx}-cZ_{nx}) \right],$$

\begin{algorithm}[!h]
\caption{Computation of $V_x^L(\alpha, \beta; \tilde{T})$ and $V_x^U(\alpha, \beta; \tilde{T})$}
\label{alg1}
\begin{algorithmic}
\REQUIRE $\gamma$, $\alpha_{0x}$, $\beta_{0x}$, $\xi_x$, $c$, and $N$
\FOR{$i = 0, ..., \tilde{T}+M$}
  \FOR{$j=\max\{0, \tilde{T}+1-\alpha\}, ..., \tilde{T}+M-\alpha$}
  	\STATE{Let $\alpha = \alpha_{0x} + i$, and $\beta = \beta_{0x}+j$.}
	\STATE{Let $V^L_x(\alpha, \beta; \tilde{T})=\frac {E[\min(M, L_{1x})]}{(1-\gamma)(1-\gamma \xi_x)} \max \left\{0, \frac{\alpha}{\alpha+\beta}-c\right\}$ and $V^U(\alpha, \beta; \tilde{T}) = \frac {E[\min(M, L_{1x})]}{(1-\gamma)(1-\gamma \xi_x)}$.}
  \ENDFOR
\ENDFOR
 \FOR{$i=N_{ttl}, ..., 0$}
	\FOR{$j=N_{ttl}-\alpha, ..., 0$}
	\STATE{Let $\alpha = \alpha_{0x}+i$, $\beta = \beta_{0x}+j$, and $\mu = \frac{\alpha}{\alpha+\beta}.$}
	\STATE{ Let
	$V^L_x(\alpha, \beta; \tilde{T})=\max_{0 \le u \le M } \left\{(\mu- c)E(\min (u, L_{1x})]+\gamma E [V^L_x(\alpha_{1x},\beta_{1x}; N_{ttl}) | U_{1x}=u] \right\},$
	$V^U_x(\alpha, \beta; \tilde{T})=\max_{0 \le u \le M } \left\{(\mu- c)E(\min (u, L_{1x})]+\gamma E [V^U_x(\alpha_{1x},\beta_{1x}; N_{ttl}) | U_{1x}=u] \right\},$}
	\ENDFOR
  \ENDFOR
\end{algorithmic}
\end{algorithm}
where a policy $\pi(x)$ is a sequence of functions, $(\pi_1(x),  \pi_2(x), ..., \pi_N(x)),$ associated with category $x$ and $\Pi(x)$ is the set of all $\pi(x)$. Each $\pi_n(x): (\{0, 1..., M\} \times \mathbb{N} \times \{0, 1..., M\})^{n-1} \mapsto \{0, 1, ..., M\}$ maps the single-category history, $\{U_{\ell x}, L_{\ell x}, Y_{\ell x}: \ell \le n-1\}$, into actions $U_{nx}$ for category $x$.

We can convert each sub-problem from a problem with a random finite time horizon to one with an infinite time horizon as follows.
% Recall $N$ is the total number of visits until the user's horizon $T$ elapses and $L_{nx}$ is the number of queued items from category $x$ between two visits. 
First, $N$ follows a geometric distribution with parameter $1-\gamma$, where $\gamma=\frac{s}{s+r}$.  Then,
\begin{eqnarray*}
E^{\pi(x)} \left[\sum^N_{n=1}(Y_{nx}-cZ_{nx}) \right]
=\gamma E^{\pi(x)} \left[\sum^\infty_{n=1} \gamma^{n-1} (Y_{nx}-cZ_{nx}) \right].
\end{eqnarray*}

We now solve this MDP using stochastic dynamic programming. We define the value function 
\begin{eqnarray}
V_x(\alpha, \beta) = \sup_{\pi(x) \in \Pi(x)} E^{\pi(x)} \left[\sum^\infty_{n=1}\gamma^{n-1}(Y_{nx}-cZ_{nx}) \Big| \alpha_{0x}=\alpha, \beta_{0x}=\beta \right].
\label{eq:filtering_valueFunc}
\end{eqnarray}

This is a two-dimensional dynamic problem, and we can write its Bellman equation as:
\begin{eqnarray} 
\begin{split}
&V_x(\alpha, \beta) = \max_{u \in \{0,1, ..., M\}}Q(\alpha, \beta, u), 
\label{eq:bellmanEq}
\end{split}
\end{eqnarray}
where $Q(\alpha, \beta, u)$ is the expected reward when we forward at most $u$ items and behave optimally afterwards:
\begin{eqnarray}
\begin{split}
Q(\alpha, \beta, u)
=& E[Y_{1x} - c\min (u, L_{1x})+\gamma V_x(\alpha_{1x},\beta_{1x}) | U_{1x}=u, \alpha_{0x}=\alpha, \beta_{0x}=\beta]\\
=&(\alpha /(\alpha+\beta)-c)  E[\min (u, L_{1x})] + \gamma E[V_x(\alpha_{1x}, \beta_{1x}) | U_{1x}=u, \alpha_{0x}=\alpha, \beta_{0x}=\beta].
\label{eq:q_factor}
\end{split}
\end{eqnarray}

The first in equation~\eqref{eq:q_factor} is the immediate reward for the action.  To compute the second term, we use that $L_{nx}$ is geometric with parameter $\xi_x= \frac{s}{\lambda_x+s}$, and so $E[\min(u, L_{1x})]=\frac{1-\xi_x}{\xi_x} \left[1-(1-\xi_x)^u \right]$. The third term in equation~\eqref{eq:q_factor} specifies the expected future reward for forwarding at most $u$ items,
\begin{eqnarray*}
&&E[V_x(\alpha_{1x}, \beta_{1x})|U_{1x}=u, \alpha_{0x}=\alpha, \beta_{0x}=\beta] \\
&=&\sum^{u}_{i=0}\sum_{k=0}^{i}P\left(Y_{1x}=k|Z_{1x}=i, \alpha_{0x}=\alpha, \beta_{0x}=\beta \right)P(Z_{1x}=i|U_{1x}=u) V_x(\alpha+k,\beta+i-k).
%&=&
%\begin{cases}
%\xi_x V_x(\alpha, \beta) + \sum_{i=1}^u \sum_{k=0}^{i}P\left(Y_{1x}=k|Z_{1x}=i, \alpha_{0x}=\alpha, \beta_{0x}=\beta \right) \\
%~~~~~~~~~~~~~~~~~~~~~~~~~~~~~~~~~~~~~~~~*P(Z_{1x}=i|U_{1x}=u) V_x(\alpha+k,\beta+i-k) ~~~~\text{  if } u \ge 1, \\
%V_x(\alpha, \beta)  ~~~~~~~~~~~~~~~~~~~~~~~~~~~~~~~~~~~~~~~~~~~~~~~~~~~~~~~~~~~~~~~~~~~~~~~~~~~~~~~~~~~~~~~~~~~~~~~~~~~~~~\text{  if } u=0.
%\end{cases}
\end{eqnarray*}

To compute this, 
one can show that
$P\left(Y_{1x}=k ~|~ Z_{1x}=i, \alpha_{0x}=\alpha, \beta_{0x}=\beta \right)=\binom {i}{k} \frac{B(\alpha+k, \beta+i-k)}{B(\alpha, \beta)}$,
%\begin{eqnarray*}
%&&P\left(Y_{1x}=k ~|~ Z_{1x}=i, \alpha_{0x}=\alpha, \beta_{0x}=\beta \right)
%= \int_{\theta=0}^1 P(Y_{1x}=k|Z_{1x}=i, \theta) dP(\theta) \\
%&=&\int_{\theta=0}^1 \binom {i} {k} \theta^{k} (1-\theta)^{i-k}\frac{1}{B(\alpha, \beta)} \theta^{\alpha-1}(1-\theta)^{\beta-1} d\theta
%=\binom {i}{k} \frac{B(\alpha+k, \beta+i-k)}{B(\alpha, \beta)},
%\end{eqnarray*}
where $B(x, y)$ is the beta function. 
% To compute the remaining terms, we must determine the distribution of $L_{nx}$, the number of queued items from category $x$ between two visits. For each visit $n \leq N$, let $\Delta_{n}$ be the elapsed time between the $(n-1)^{th}$ and $n^{th}$ visits of the given user. Then, $L_{nx}|\Delta_{n}$ is an independent and identically distributed Poisson random variable with mean $\lambda_x \cdot \Delta_{n}$. Through some algebra, one can then show that 
%\begin{eqnarray*}
%P(L_{nx}=\ell) 
%&=& \int^\infty_{t=0} P(L_{nx}=\ell |\Delta_{n}=t) s e^{-s t} dt 
%=\int^\infty_{t=0} \frac{(\lambda_x t)^\ell e^{-\lambda_x t}}{\ell!} s e^{-s t} dt \\
%&=&\frac{\lambda_x^\ell}{\ell !}s \int^\infty_{t=0} t^\ell e^{-t(\lambda_x+s)}dt
%=\frac{\lambda_x^\ell}{\ell !} \int^\infty_{z=0} \left( \frac{z}{\lambda_x+s}\right)^\ell e^{-z} \frac{dz}{\lambda_x+s}  \\
%&=& \frac{\lambda_x^\ell}{\ell !} s [(\lambda_x+s)^{-(\ell +1)}] \ell ! 
%= \left(\frac{\lambda_x}{\lambda_x+s}\right)^\ell \left(\frac{s}{\lambda_x+s}\right).
%\end{eqnarray*}
The distribution $Z_{1x}=\min(U_{1x},L_{1x})|U_{1x}$ is given by:
\begin{align*}
P(Z_{1x} = i|U_{1x}=u)=
\begin{cases}
(1-\xi_x)^i\xi_x & \text{if } 0 \le i < u, \\
(1-\xi_x)^{u} & \text{if } i = u.
\end{cases}
\end{align*}

%Also using that $L_{nx}$ is geometric with parameter $\xi_x$, we have $E[\min(u, L_{1x})]=\frac{1-\xi_x}{\xi_x} \left[1-(1-\xi_x)^u \right]$.
%\begin{eqnarray*}
%E\left[\min(u, L_{1x}) \right]
%&=& \sum_{k=0}^{u-1} k (1-\xi_x)^k \xi_x + u \left[1-\sum^{u-1}_{k=0}(1-\xi_x)^k\xi_k \right] \\
%&=&\xi_x (1-\xi_x) \frac{d}{d\xi_x} \left[-\sum_{k=1}^{u-1}(1-\xi_x)^k \right] + u (1-\xi_x)^{u} \\
%&=&-\xi_x(1-\xi_x) \left[\frac{u(1-\xi_x)^{u-1}\xi_x-[1-(1-\xi_x)^u]}{\xi_x^2} \right] + u(1-\xi_x)^{u}\\
%&=&\frac{1-\xi_x}{\xi_x} \left[1-(1-\xi_x)^u \right].
%\end{eqnarray*}

With these expressions, we can use backward induction, illustrated in Algorithm~\ref{alg1}, to solve the dynamic program in equation \eqref{eq:filtering_valueFunc} by considering a truncated time-horizon problem terminated at $\tilde{T}$, and averaging an upper bound on the value function, $V^L_x(\alpha, \beta; \tilde{T})$ and a lower bound, $V^U_x(\alpha, \beta; \tilde{T})$. 

\section{Mathematical Model with a Constraint on Items Forwarded} \label{sec:LagrangeRelaxation}
In Section \ref{sec:filtering_model}, we assumed that the system knows the unit cost, $c$, that the user incurs for reviewing each item. In reality, we often do not know this cost. In this section, we instead assume that the number of items forwarded in each step is constrained, $\sum^k_{x=1}U_{nx}\le M\ \forall n$. Our objective is to maximize the expected number of relevant items forwarded, subject to this constraint:
\begin{eqnarray}
&\sup_{\tilde{\pi} \in \tilde{\Pi}}E^{\tilde{\pi}} \left [ \sum^N_{n=1}\sum^k_{x=1}Y_{nx} \right], %\\
%&\text{s.t.}~~~~ \sum_{x=1}^k Z_{nx} \leq M ~~\forall~ n.
\label{eq:rankingDP}
\end{eqnarray}
where $\tilde{\Pi}=\{\pi \in \Pi: \sum^k_{x=1}U_{nx}\le M\ \forall n\}$ and $\tilde{\pi}$ is a policy in $\tilde{\Pi}$ that satisfies the constraint. 

In contrast with the previous problem, computation in this problem scales exponentially in $k$ due to ``curse of dimensionality'', because we can no longer decompose equation \eqref{eq:rankingDP} into multiple tractable sub-problems \citep{PowellADP2007}. Instead, we consider a Lagrangian relaxation, following developments \cite{HuFrazierXie2014,XieFrazier2013b}, that provides a computationally tractable upper bound on the value of equation \eqref{eq:rankingDP}, and which motivates an index-based heuristic policy below in Section~\ref{sec:MDPIF}.

Let $\boldsymbol \alpha_{0} = (\alpha_{01}, ..., \alpha_{0k})$ and $\boldsymbol \beta_{0} = (\beta_{01}, ..., \beta_{0k})$. Let $\boldsymbol \nu = (\nu_1, ..., \nu_N)$ be a vector of Lagrange multipliers, with each $\nu_n \ge 0$ denoting a unit cost (or penalty) when we violate the constraint, $\sum_{x=1}^k Z_{nx} \leq M$, at step $n$. We can then write the Lagrangian relaxation of \eqref{eq:rankingDP} as
\begin{eqnarray} \label{eq:lagrangianRelaxation}
V^{\boldsymbol \nu}(\boldsymbol \alpha_0, \boldsymbol \beta_0) &=&\sup_{\pi \in \Pi} E^\pi \left [ \sum^N_{n=1} \left[ \sum^k_{x=1} Y_{nx}- \nu_n \left(\sum_{x=1}^k Z_{nx} - M \right) \right] \right]\\ %=\sup_{\pi \in \Pi} E^\pi \left [ \sum^k_{x=1} \sum^N_{n=1}(Y_{nx}- \nu_n Z_{nx}) \right] + M E\left(\sum^N_{n=1} \nu_n \right) \\
&=&\sum_{x=1}^k \sup_{\pi(x) \in \Pi(x)} E^{\pi(x)} \left [ \sum^N_{n=1}(Y_{nx}- \nu_n Z_{nx}) \right] + M E\left[\sum^N_{n=1} \nu_n \right]. %\\&=& \sum_{x=1}^k V^{\boldsymbol \nu}_x(\alpha_{0x}, \beta_{0x}) + M E\left(\sum^N_{n=1} \nu_n \right),
\end{eqnarray}
Because of the constraint $\sum^{k}_{x=1} Z_{nx} \le M$ for all $n$ satisfied by $\tilde{\pi}\in\tilde{\Pi}$, and the non-negativity of the Lagrange multiplier, $\boldsymbol \nu \ge \boldsymbol 0$, the value in equation \eqref{eq:lagrangianRelaxation} provides an upper bound on the value of equation \eqref{eq:rankingDP}. Given $\boldsymbol \nu \ge \boldsymbol 0$, let $V^{\boldsymbol \nu}_x(\alpha_{0x}, \beta_{0x}) = \sup_{\pi \in \Pi} E^\pi \left [ \sum^N_{n=1}(Y_{nx}- \nu_n Z_{nx}) \right]$.  Then $V^{\boldsymbol \nu}(\boldsymbol \alpha_0, \boldsymbol \beta_0)$ can be decomposed into a sum of multiple sub-problems, plus a constant, 
\begin{eqnarray}
V^{\boldsymbol \nu}(\boldsymbol \alpha_0, \boldsymbol \beta_0)
&=&\sum_{x=1}^k V^{\boldsymbol \nu}_x(\alpha_{0x}, \beta_{0x}) + M E\left[\sum^N_{n=1} \nu_n \right]
\ge  \sup_{\tilde{\pi} \in \tilde{\Pi}} E^{\tilde\pi} \left [ \sum^N_{n=1}\sum^k_{x=1}Y_{nx} \right].
\label{eq:V_nu}
\end{eqnarray}
This upper bound $V^{\boldsymbol \nu}(\boldsymbol \alpha_0, \boldsymbol \beta_0)$ is useful because it can be computed efficiently, as the sum of independent and easy-to-solve stochastic control sub-problems. Furthermore, if we consider this upper bound for the special case $\boldsymbol \nu = \nu \boldsymbol e$ with $\nu \ge 0$ and $\boldsymbol e = (1, 1, ..., 1)$, then $V_x^{\boldsymbol \nu}(\alpha_{0x}, \beta_{0x})$ recovers the total expected reward in the information filtering problem described in Section \ref{sec:filtering_model}, with a fixed unit cost, $\nu$, for forwarding each item. $V_x^{\boldsymbol \nu}(\alpha, \beta)$ can be computed efficiently using Algorithm 1. 

We can construct tighter upper bound by taking the infimum of $V^{\boldsymbol \nu}(\boldsymbol \alpha_0, \boldsymbol \beta_0)$ over sets of potential values for our Lagrange multipliers,
\begin{eqnarray}
\text{UB}({\boldsymbol \alpha}_{0}, {\boldsymbol \beta}_{0}) 
&=& \inf_{\boldsymbol \nu \ge 0, \boldsymbol \nu= \nu \boldsymbol e} V^{\boldsymbol \nu}(\boldsymbol \alpha_0, \boldsymbol \beta_0)
\ge  \inf_{\boldsymbol \nu \ge 0} V^{\boldsymbol \nu}(\boldsymbol \alpha_0, \boldsymbol \beta_0).\label{eq:upperBound}
\end{eqnarray}
Inequality \eqref{eq:upperBound} holds because $\{\boldsymbol \nu \ge \boldsymbol 0: \boldsymbol \nu = \nu \boldsymbol e\} \subseteq \{\boldsymbol \nu \ge \boldsymbol 0\}$.

We conjecture that the function $V_x^{\boldsymbol \nu}(\alpha_{0x}, \beta_{0x})$ is convex and non-increasing in $\boldsymbol \nu$. The other term, $M E\left[ \sum_{n=1}^N \nu_n\right]$ is also a convex function of $\boldsymbol \nu$, so $V^{\boldsymbol \nu}(\boldsymbol \alpha_0, \boldsymbol \beta_0)$ is conjectured to be convex. If this conjecture is true, we can use a bisection algorithm to find $\boldsymbol \nu \ge \boldsymbol 0$ in the region $\{\boldsymbol \nu \ge \boldsymbol 0: \boldsymbol \nu= \nu \boldsymbol e\}$ that minimizes 
$V^{\boldsymbol \nu}({\boldsymbol \alpha}_{0}, {\boldsymbol \beta}_{0})$, obtaining $\text{UB}(\boldsymbol \alpha_0, \boldsymbol \beta_0)$. The conjectured non-increasing property of $V_x^{\boldsymbol \nu}(\alpha_{0x}, \beta_{0x})$ induces another property of the optimal solution: given $\nu>0$, if it is optimal to forward $m$ items at $(\alpha_{0x}, \beta_{0x})$, then it should be optimal to forward at least $m$ items for all $0 \le \nu' < \nu$. 

Given $\boldsymbol \nu = \nu \boldsymbol e$, let $\pi_{\nu}^*(x)$ be the optimal policy for $V_x^{\boldsymbol \nu}(\cdot, \cdot)$, and $U^*(\alpha, \beta) \in \pi_{\nu}^*(x)$ be the optimal decision at state $(\alpha, \beta)$. For each $0 \le u \le M$, we define $\nu^*(u, \alpha, \beta)$ to be the largest reward achieved for forwarding (or the largest cost a user would pay to view) at least $u$ number of items at state $(\alpha, \beta)$, 
\begin{eqnarray} 
%\begin{split}
\label{eq:cstar}
\nu^*(u, \alpha, \beta) 
&=&\sup_{\nu \ge 0} \left\{\nu: U_{\nu}^*(\alpha, \beta) \ge u \text{ with } U^*_{\nu}(\alpha, \beta) \in \pi_{\nu}^*(x) \right\}\\
&=& \sup_{\nu \ge 0} \left\{\nu: V^{\boldsymbol \nu}(\alpha, \beta) = \max_{u'\ge u} Q^{\nu}(\alpha, \beta, u') \right\}.
%\end{split}
\end{eqnarray}
%where 
%$Q^\nu(\alpha, \beta, u) = Y_{1x}- \nu \min \{u, L_{1x}\} + E\left[V^\nu(\alpha_{1x}, \beta_{1x}) | U_{1x}=u, \alpha_{0x}=\alpha, \beta_{0x}=\beta \right].$
In the special case $M=1$ and $L_{nx} \ge 1$ for all $n$ and $x$, the index $\nu^*(1, \alpha, \beta)$ actually corresponds to Gittin's index for a two-armed bandit \citep{Gittins79, Wh80}, where one arm gives the known value of not forwarding $(0)$, and the other arm gives the unknown value of forwarding in a single-category problem.
% . The MAB problem involves $n$ independent bandit processes in a Markovian framework and studies how to sequentially choose an arm at each step to maximize the total reward. In 1970s, Gittins and Jones demonstrated that the optimal solution is an index policy, in which one operates the arm with the highest Gittin's index, which measures the largest total reward one can achieve if the arm is chosen to operate at the current state.
\begin{figure}[ht]
\begin{center}
\begin{subfigure}[b]{.48\linewidth}
\centering
\includegraphics[height=4.5cm, width=6cm]{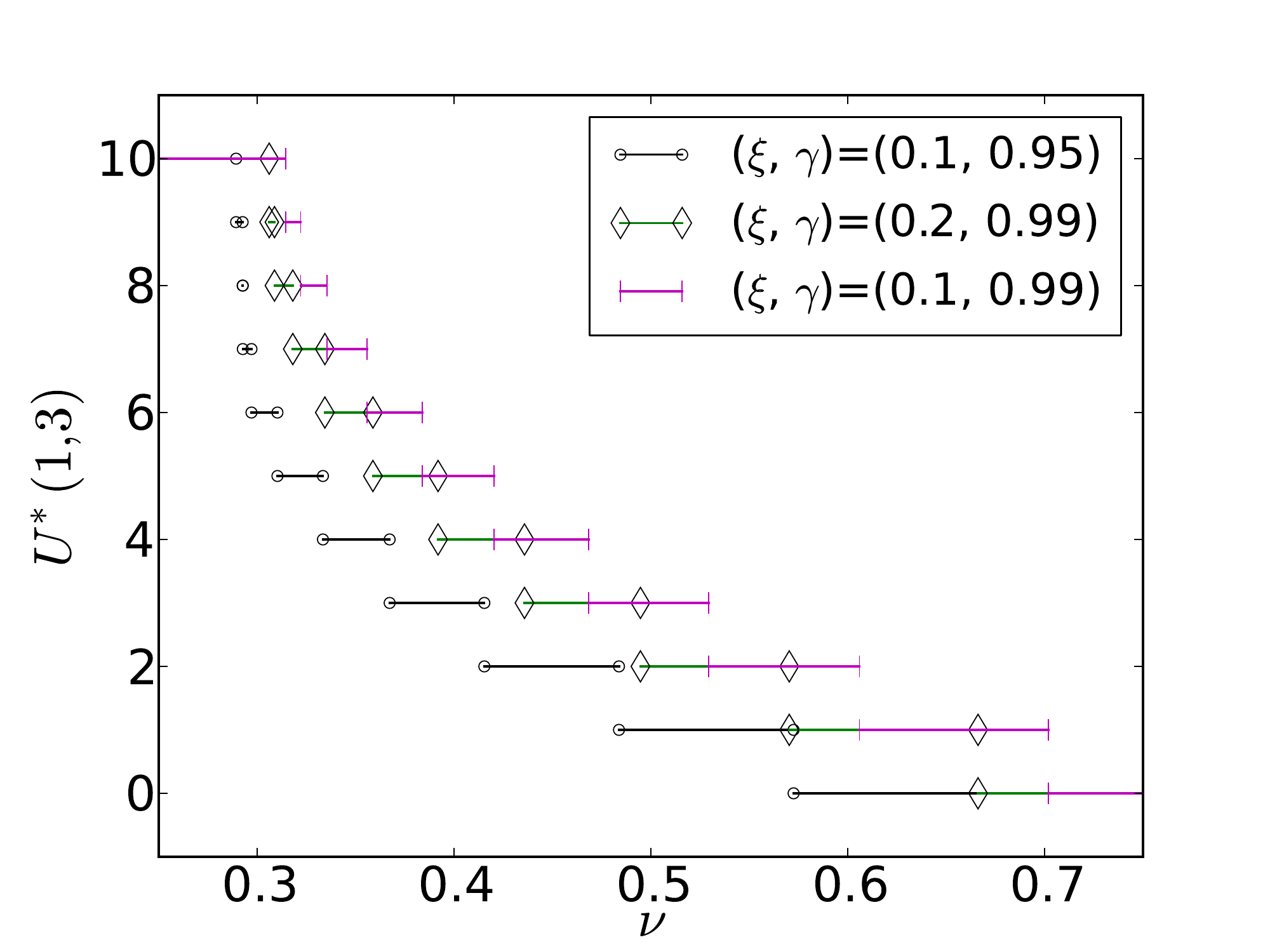}
\caption{Plot of $U^*(\alpha, \beta)$ vs. $\nu$ at state $(\alpha, \beta) = (1, 3)$ for different values of $\xi$ and $\gamma$.}
\end{subfigure}
\quad
\begin{subfigure}[b]{.48\linewidth}
\centering
\includegraphics[height=4.5cm,width=6cm]{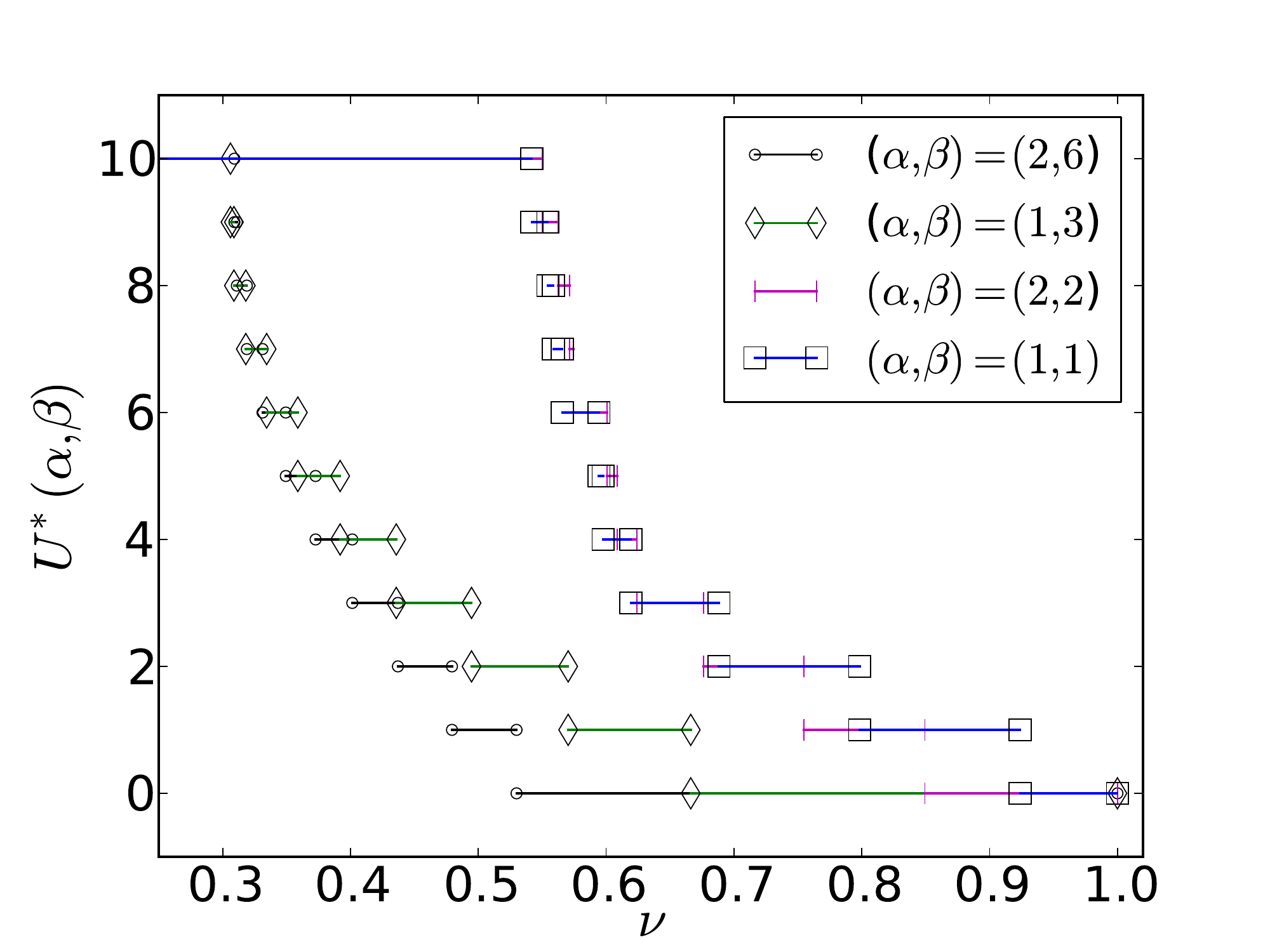}
\caption{Plot of $U^*(\alpha, \beta)$ vs. $\nu$ at $\xi=0.2$ and $\gamma=0.99$ for different pairs of $(\alpha, \beta)$. }
\end{subfigure}
\end{center}
\caption{Plots of optimal decision $U^*(\alpha, \beta)$ against Lagrange multiplier $\boldsymbol \nu = \nu \boldsymbol e$ for various values of $(\alpha, \beta, \xi, \gamma)$ with $M=10$. In each case, we solve $V_x^\nu(\alpha, \beta)$ via Algorithm \ref{alg1} to find $U^*(\alpha, \beta)$ for each $\nu \in [0, 1]$. Here, $\nu^*(u, \alpha, \beta)$ is $\nu$ at the right-most endpoint of the interval where $U^*(\alpha, \beta) \ge u$. }\label{fig:cstar_plot}
\end{figure}
Figure \ref{fig:cstar_plot} plots the optimal decision $U^*(\alpha, \beta)$ against Lagrange multiplier ${\bf \nu} = \nu {\bf e}$ for different values of $\alpha$, $\beta$, $\xi$, and $\gamma$. In both plots, we see that $U^*(\alpha, \beta)$ is non-increasing in $\nu$. The observation confirms the conjectured statement that if it is optimal to forward $m$ items at $\nu$, then it is optimal to forward at least $m$ items for all $\nu'<\nu$. Furthermore, $\nu^*(u, \alpha, \beta)$ is $\nu$ at the right-most endpoint of intervals where $U^*(\alpha, \beta) \ge u$. In Figure \ref{fig:cstar_plot}(a), we vary $\xi$ and $\gamma$ while fixing $(\alpha, \beta)$. Increasing the discount factor, $\gamma$, increases $\nu^*(u, \alpha, \beta)$ because a user spends longer in the system. In contract, as $\xi$ increases, $\nu^*(u, \alpha, \beta)$ becomes smaller because on average there are fewer items queued in the system. Figure \ref{fig:cstar_plot}(b) illustrates that $\nu^*(u, \alpha', \beta') \ge \nu^*(u, \alpha, \beta)$ in the case when $\frac{\alpha'}{\alpha'+\beta'} = \frac{\alpha}{\alpha+\beta}$ but $\alpha'+\beta' < \alpha + \beta$. This reflects that exploration provides the largest value when observation size is small, and uncertainty is high. 

\subsection{MDP-based Information Filtering (MDP-IF) Policy} \label{sec:MDPIF}
In this section, we propose a heuristic index-based policy for ranking items queued at step $n$, disregarding whether the user cost is known. We call our policy the MDP-based Information Filtering Policy, abbreviated {\it MDP-IF}. We then discuss scenarios where the proposed policy is Bayes-optimal.
\begin{figure}[ht]
\begin{center}
\begin{subfigure}[b]{.48\linewidth}
\centering
\includegraphics[height=4.5cm, width=7.8cm]{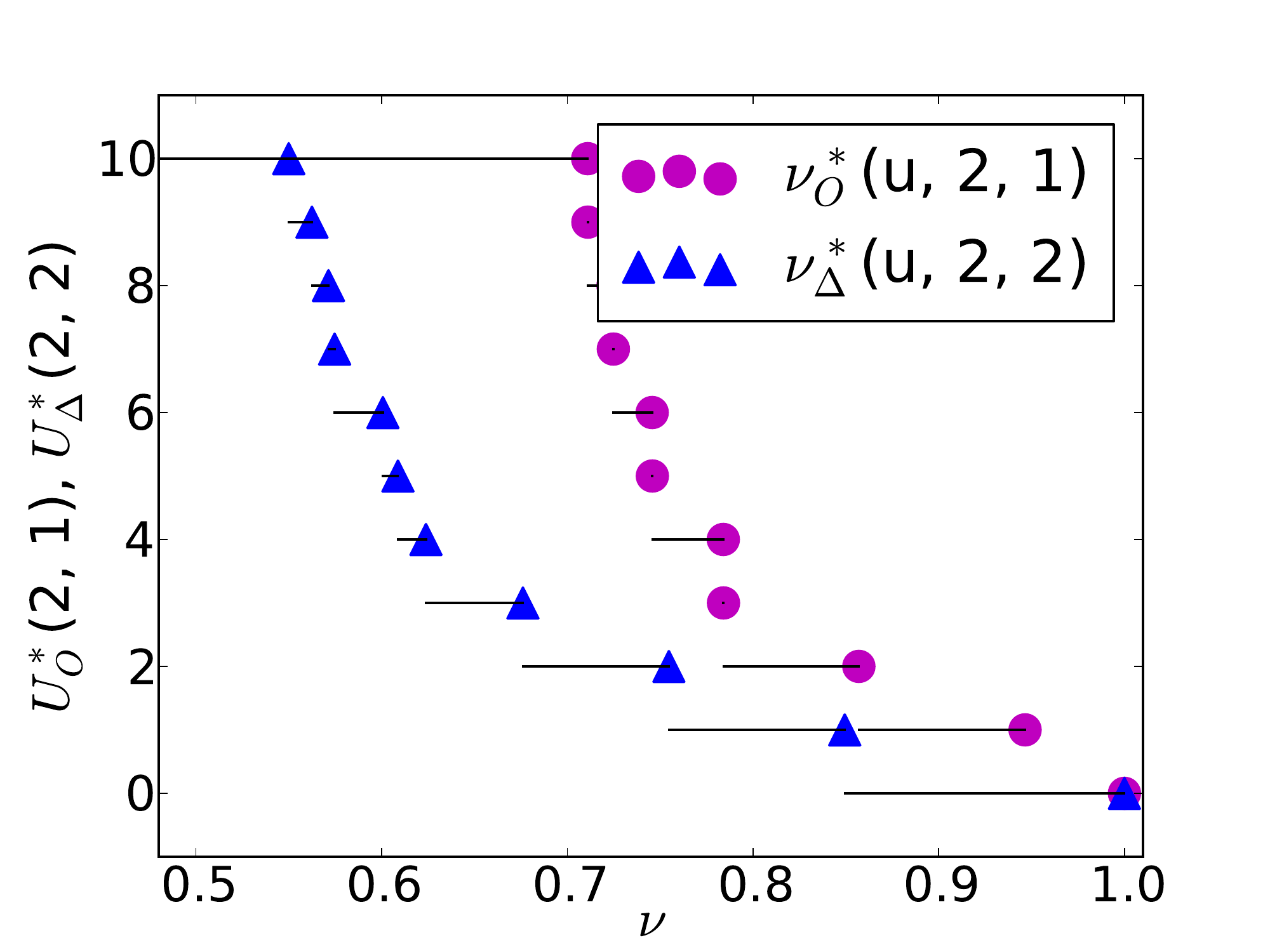}
\caption{Plot of $U^*_{O}(2, 1)$ and $U^*_{\Delta}(2, 2)$ against $\nu$.}
\end{subfigure}
\quad
\begin{subfigure}[b]{.48\linewidth}
\centering
\includegraphics[height=4.5cm, width=7.8cm]{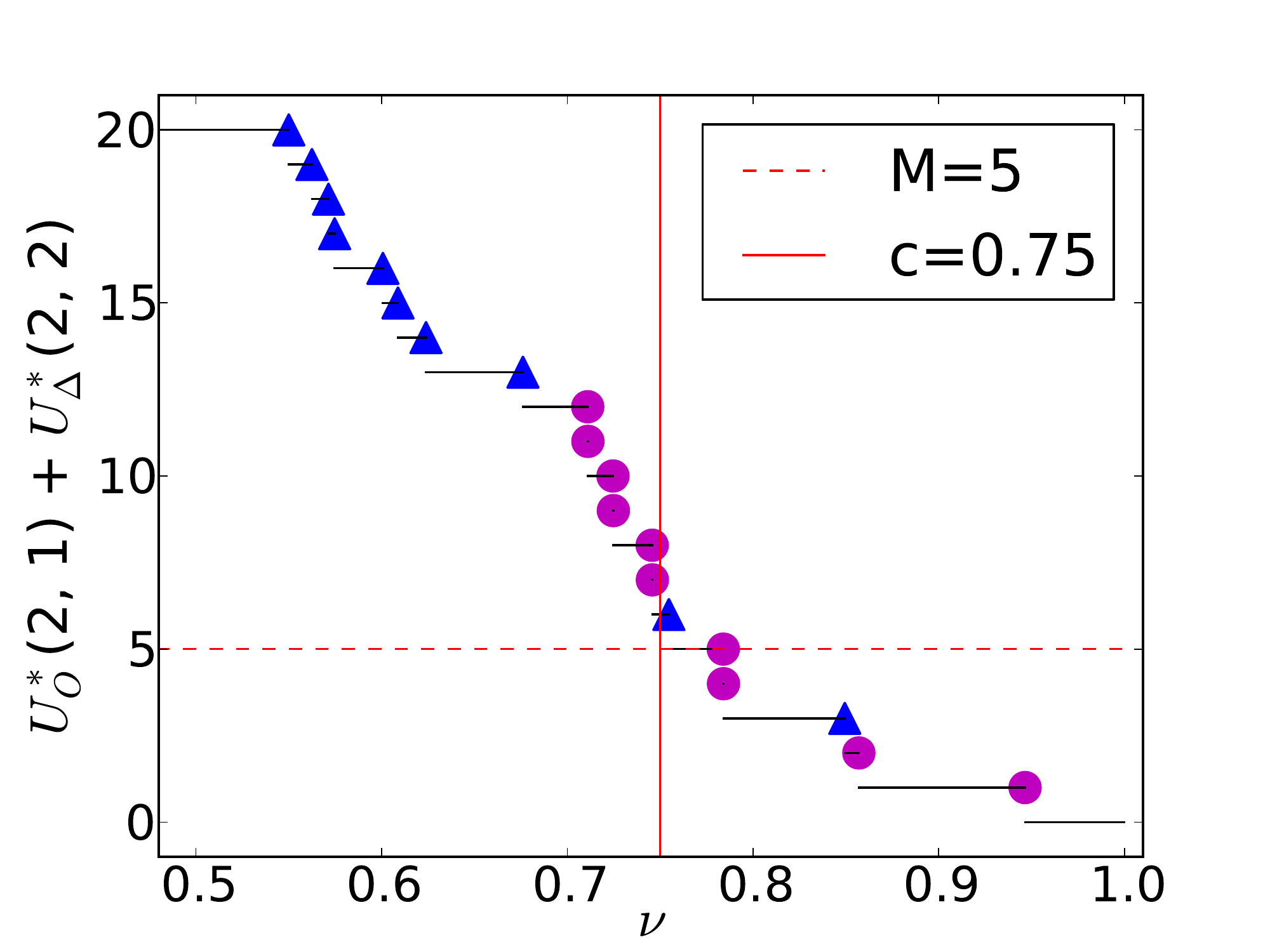}
\caption{Plot of $U^*_{O}(2, 1) + U^*_{\Delta}(2, 2)$ against $\nu$}
\end{subfigure}
\end{center}
\caption{There are two categories: category $O$ at state (2, 1) and category $\Delta$ at state (2, 2). Figure (a) shows optimal decision $U^*_{O}(2, 1)$, $U^*_{\Delta}(2, 2)$ against lagrangian multiplier $\nu$, and also plots $\nu^*_{O}(u, 2, 1)$ (denoted in purple circles) and $\nu^*_{\Delta}(u, 2, 2)$ (denoted in blue triangles). Figure (b) plots $U^*_{O}(2, 1) + U^*_{\Delta}(2, 2)$ vs. $\nu$, with ranked list of $\nu^*_{O}(u, 2, 1)$ and $\nu^*_{\Delta}(u, 2, 2)$ among all $u \in \{0, 1, ..., 10\}$. When $M=5$, the rank list would be: $\{O, O, \Delta, O, O \}$. When the user cost is 0.75, the rank list would be the six right-most items, $\{O, O, \Delta, O, O, \Delta\}$.}\label{fig:cstar_decision}
\end{figure}
At each step $n$, we first compute $\nu^*(u, \alpha_{nx}, \beta_{nx})$ for all possible $u \le M$ in each category $x$, then rank items from the categories based on the computed $\nu^*(u, \alpha_{nx}, \beta_{nx})$, among all categories and all $u \le M$. $\nu^*(u, \alpha_{nx}, \beta_{nx})$ reflects the highest reward one can achieve if we choose to forward at least $u$ items from the category $x$ at the current state, $(\alpha_{nx}, \beta_{nx})$. In the situation where we have a constraint that $\sum_{x=1}^kZ_{nx} \le M$, we would forward items from the ranked list until we exhaust $M$ slots. This heuristic index policy is exactly the optimal policy corresponding to the Gittins index in a conventional multi-armed bandit problem, when there is at least one item queued in each category per step and $M=1$. On the other hand, if a user cost $c$ is specified, then we would forward items from the ranked list with $\nu^*(u, \alpha_{nx}, \beta_{nx}) \ge c$. In a general case with both $c$ and $M$ specified, we follow the same procedure with a stop criteria if any condition is violated. 

Figure \ref{fig:cstar_decision} demonstrates how this index policy works in a two-category problem (category ``O'' and category ``$\Delta$'').  Category ``O'' is at state (2, 1) and  category ``$\Delta$'' is at state (2, 2). Figure (a) plots optimal decisions $U^*_{O}(2, 1)$ and $U^*_{\Delta}(2, 2)$ as a function of $\nu$, and identifies their $\nu^*_{O}(u, 2, 1)$ and $\nu^*_{\Delta}(2, 2)$ values. Then, Figure (b) plots $U^*_{O}(2, 1) + U^*_{\Delta}(2, 2)$ against $\nu$, with a ranked list of $\nu^*_{O}(u, 2, 1)$ and $\nu^*_{\Delta}(u, 2, 2)$ among all $u \in \{0, 1, ..., 10\}$. When $M=5$, the ranked list would be items below the dashed line: $\{O, O, \Delta, O, O \}$. When the user cost is 0.75, the ranked list would be the six items on the right of the vertical line (defined by $c=0.75$): $\{O, O, \Delta, O, O, \Delta\}$.
\section{Experimental Results}
\label{sec:experiments}
In this section, we show numerical results of Monte Carlo simulations in four different parameter settings to illustrate how this heuristic index-based policy performs compared to other competing methods, including the pure exploitation and upper confidence bound (UCB) policies. At each step $n+1$, pure exploitation ranks items based on posterior means of user preference across categories, while UCB ranks items by $\left(1-1/t_{n}\right)\%$ quantiles from the posterior distribution on user preferences, with $t_{n}$ denoting the total number of items shown by $n$. 

\begin{figure}[ht]
\begin{center}
\begin{subfigure}[b]{.225\linewidth}
\centering
\includegraphics[height=3.5cm, width=3.7cm]{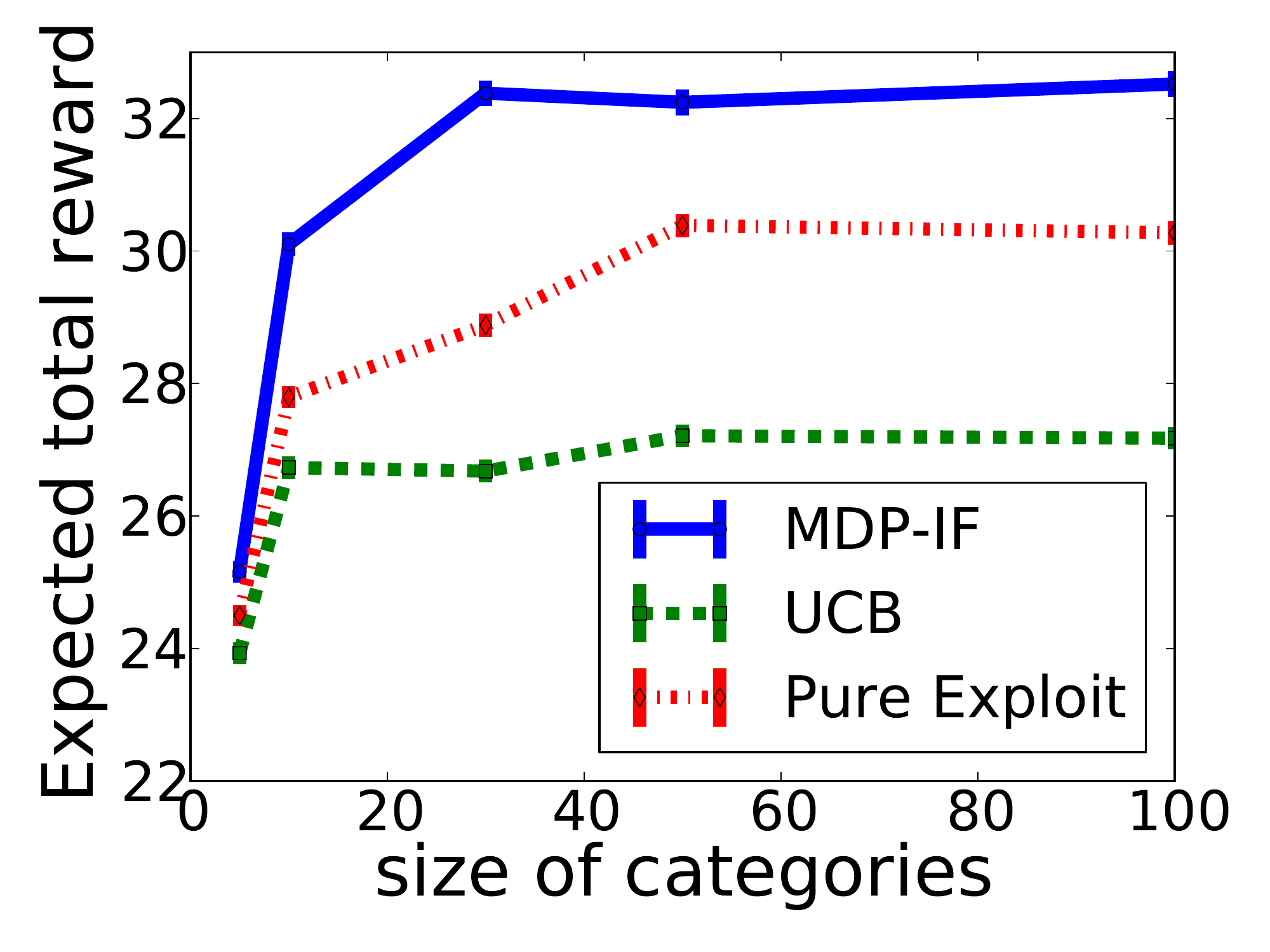}
\caption{$M=5$, $c=.49$, $\mathbf{\gamma=0.95}$, $\alpha_{0x}=1$, $\beta_{0x}=1$, $\xi_x=0.1$ $\forall x$}
\end{subfigure}
\quad
\begin{subfigure}[b]{.225\linewidth}
\centering
\includegraphics[height=3.5cm, width=3.8cm]{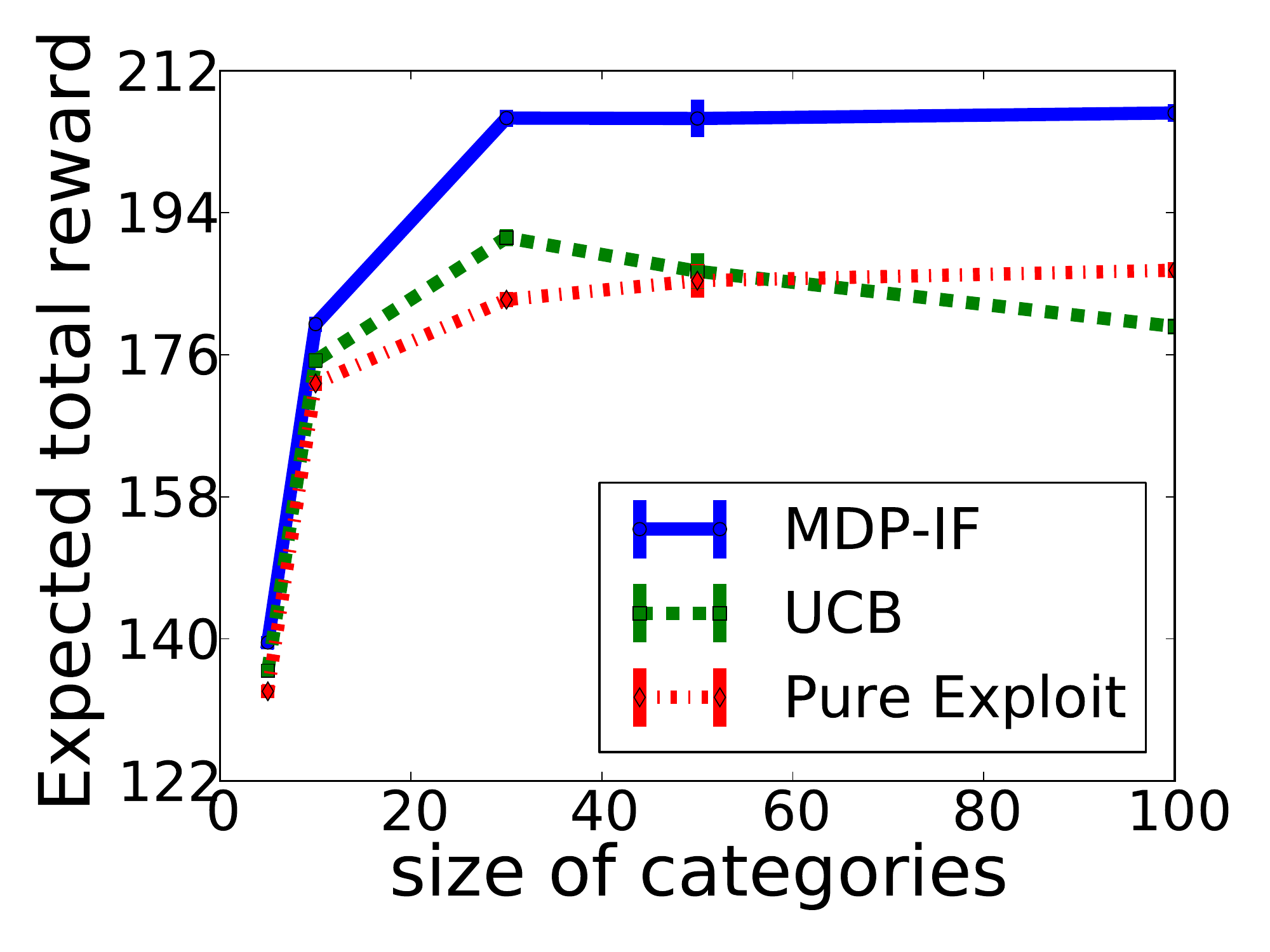}
\caption{$M=5$, $c=.49$, $\mathbf{\gamma=0.99}$, $\alpha_{0x}=1$, $\beta_{0x}=1$, $\xi_x=0.1$ $\forall x$}
\end{subfigure}
\quad
\begin{subfigure}[b]{.225\linewidth}
\centering
\includegraphics[height=3.5cm, width=3.7cm]{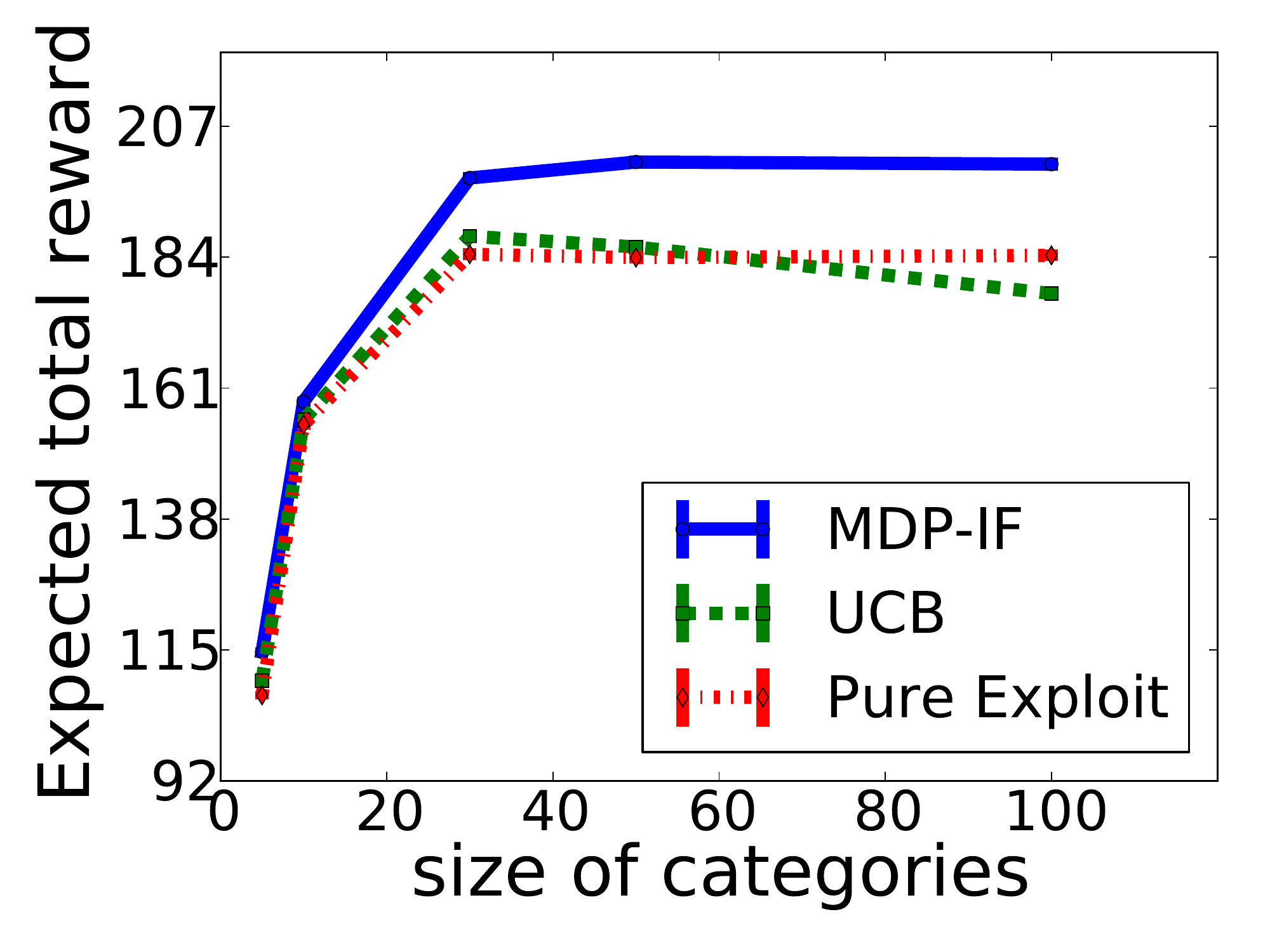}
\caption{$M=5$, $c=.49$, $\gamma=0.99$, $\alpha_{0x}=1$, $\beta_{0x}=1$, $\mathbf{\xi_x=0.2}$ $\forall x$}
\end{subfigure}
\quad
\begin{subfigure}[b]{.225\linewidth}
\centering
\includegraphics[height=3.5cm, width=3.8cm]{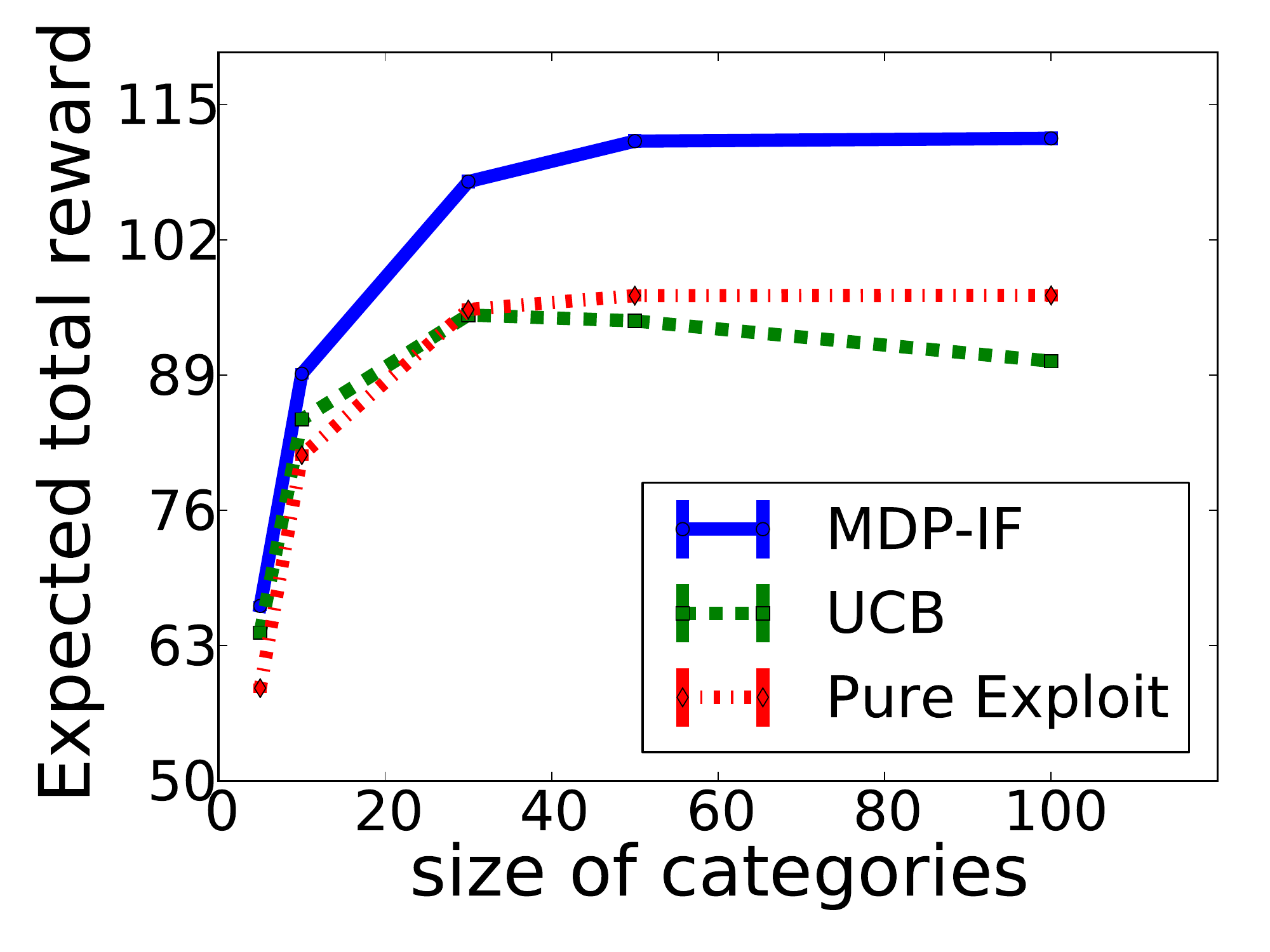}
\caption{$M=5$, $c=.49$, $\gamma=0.99$, $\mathbf{\alpha_{0x}=5}$, $\mathbf{\beta_{0x}=5}$, $\xi_x=0.1$ $\forall x$}
\end{subfigure}
\end{center}
\caption{Each sub-figure shows plots of expected total reward with 95\% confidence intervals under each policy (MDP-IF in solid blue line, UCB in red dashed line, Pure Exploit in green dotted line) against category size in a given parameter setting specified in its sub-caption. $50,000$ users are simulated in each scenario.}\label{fig:sim_results}
\end{figure} 
 
We run experiments for four different scenarios, each with a chosen set of parameters, including $\gamma$ and $\{\alpha_{0x}, \beta_{0x}, \xi_x\}_{x \in \{1, ..., k\}}$. Each scenarios consists of simulations for a range of $k \in \{5, 10, 30, 50, 100\}$, for us to understand how each policy behaves as category size, $k$, varies. Specific parameters setting are described in sub-figure captions in Figure~\ref{fig:sim_results}. Here, $\gamma$ is chosen based on the estimated parameter of the empirical distribution of user visits in astro-ph.GA and astro-ph.CO from the arXiv.org dataset in 2009-2010. Similarly, $\xi_x$ is chosen based on the empirical distributions. For convenience, $\alpha_{0x}$ and $\beta_{0x}$ is chosen to be 1 for three scenarios, and $5$ for the last scenario to understand how the policy performs when we are more certain about user preferences. For constraints, we set $M=5$, the maximum number of items forward, and unit cost as $c=0.49$, which is motivated by our conjecture that exploration adds the most benefit when cost is near the initial mean of $\theta_x$, which is $0.5$ in our test cases. 

With the chosen set of parameters in each scenario, we simulate a large set of users, with user visits, items arrivals between two visits, and user feedbacks on forwarded items generated according to the model formulated in Section~\ref{sec:filtering_model}. At each visit, each policy (MDP-IF, UCB, or pure exploitation) decides how many and which items to forward. The reward is then collected to compute the expected total reward for each policy.
  
Each line in Figure~\ref{fig:sim_results} shows the total expected reward with 95\% confidence intervals against category size, $k$, for a given parameter setting. In all scenarios, the MDP-IF policy outperforms both UCB and pure exploitation, with magnified improvement for a larger category size and, mostly, a non-decreasing relationship with $k$. For some larger $k$, the improvement from the MDP-IF policy widens as $\gamma$ increases (comparing Figure~\ref{fig:sim_results}(a) and (b)) or $\xi_x$ shrinks (comparing Figure~\ref{fig:sim_results}(b) and (c)), or the prior user preference has less variance (comparing Figure~\ref{fig:sim_results}(b) and (d)).
\section{Conclusion}
We consider the information filtering problem where we face a voluminous stream of items and need to decide sequentially on which batch of items to forward to a user so that the total reward, the number of relevant items shown minus the cost of forwarded items, is maximized. With a focus on limited historical data, we formulate the problem as a Markov Decision Process in a Bayesian setting, and then provide a computationally tractable algorithm that is Bayes-optimal. In a setting where the total number of item shown is constrained, we consider a Lagrangian relaxation of the problem and provide an index-based policy that ranks items. As shown in the numerical section, our index-based policy outperforms UCB and the pure exploitation policy, and provides magnified benefits in many settings. 

\section*{Acknowledgements}
Authors were supported by NSF IIS-1247696 and IIS-1247696.  Peter Frazier was also supported by NSF CAREER CMMI-1254298, AFOSR YIP FA9550-11-1-0083, and AFOSR FA9550-12-1-0200.

\bibliographystyle{ormsv080}
\bibliography{xzhao}
\end{document}